\documentclass[11pt]{article}
\usepackage{enumitem}
\usepackage[preprint]{acl}
\usepackage{times}
\usepackage{latexsym}
\usepackage{svg}
\usepackage[T1]{fontenc}

\usepackage[utf8]{inputenc}
\usepackage{microtype}
\usepackage{graphicx}
\usepackage{booktabs}
\usepackage{pifont}
\usepackage{verbatim}
\usepackage{fancyvrb}
\usepackage{fvextra}
\usepackage[most]{tcolorbox}
\newtcolorbox{promptbox}[1]{%
  enhanced, breakable,
  colback=yellow!7,
  colframe=yellow!55!black,
  fonttitle=\bfseries,
  coltitle=black,
  title=#1,
  arc=2mm, boxrule=0.6pt,
  left=6pt, right=6pt, top=4pt, bottom=4pt,
  attach boxed title to top left={xshift=4mm,yshift=-2.2mm},
  boxed title style={
    colback=yellow!20, colframe=yellow!55!black,
    arc=1mm, boxrule=0.4pt
  },
}
\title{Iteration Without Elaboration: A Simple ReAct Architecture Suffices for Text-to-SQL Generation}

\author{
 \textbf{Jian Lu\textsuperscript{1}},
 \textbf{Haiwei Yu\textsuperscript{1}},
 \textbf{Raymond M Xiong\textsuperscript{1}},
 \textbf{Anru Zhang\textsuperscript{1}\textsuperscript{$\dagger$}},
 \textbf{Danyang Zhuo\textsuperscript{1}\textsuperscript{$\dagger$}}
\\[4pt]
 \textsuperscript{1}Duke University
\\
 \textsuperscript{$\dagger$}Equal supervision and corresponding authors.
\\[4pt]
 \text{
  \{jian.lu, haiwei.yu, raymond.xiong, anru.zhang\}@duke.edu; danyang@cs.duke.edu
 }
}

\begin{document}
\maketitle
\begin{abstract}
Modern text-to-SQL systems have become increasingly elaborate, relying on schema-linking modules, retrieval-augmented prompting, candidate generation, and multi-stage refinement pipelines. While effective, these additions introduce substantial latency and engineering overhead. To this end, we present \textbf{ReAct-SQL}, a simple yet effective zero-shot ReAct-style framework built solely on iterative reasoning and a constrained action space defined by a typed Domain-Specific Language (DSL) of 15 relational operations, rather than free-form SQL generation. The model incrementally issues DSL calls, observes compiled-SQL execution feedback, and revises its reasoning through interaction. 
On corrected BIRD mini-dev and EHR-SQL, ReAct-SQL achieves \textbf{84.5\%} and \textbf{73.9\%} accuracy, respectively, matching substantially more elaborate baselines while running up to $8\times$ faster. Incremental ablations further show that iteration primarily improves grounding, while the DSL improves compositional reliability.
\end{abstract}

\section{Introduction}
\label{sec:intro}

Text-to-SQL---translating natural-language questions into executable queries
against a relational database---is a long-standing problem that large language
models (LLMs) have recently achieved substantial performance improvements on popular benchmarks\cite{li2023bird,yu2018spider}. Yet unlike most code-generation tasks, SQL is not a free-standing programming language: its correctness is inseparable from the database it
targets. The same question against two schemas can require completely
different queries, and a single wrong column name, misaligned join, or format
mismatch on a join key silently invalidates the answer. In this sense,
text-to-SQL is less a code-generation problem than a \emph{schema-conditioned
reasoning} problem.

Recent work spans one-shot prompting, fixed multi-stage pipelines,
feedback-driven agentic loops, and retrieval-augmented in-context
learning (\S\ref{sec:related})---each adding components for incremental
accuracy gains, but underweighting their resulting \emph{cost-latency
profile}. Latency in particular is a hard constraint that money cannot
buy its way out of: a pipeline with high latency is unusable in an
interactive setting regardless of its budget. A second deployment
friction comes from in-context retrieval: pipelines that rely on
retrieved question--SQL pairs cannot be applied to private or
enterprise databases without first labeling a corpus.

Empirically, we find that text-to-SQL failures fall into two largely orthogonal categories. The first arises when the target query is long or deeply nested: the model hallucinates non-existent tables or columns, or emits syntactically invalid SQL. The second arises when schema and table names are opaque (e.g.\ \texttt{chartevents}, \texttt{d\_icd\_diagnoses}): the model cannot reliably identify which table or column the question refers to. We refer to these as the \emph{composition problem} (producing well-formed, executable SQL) and the \emph{grounding problem} (mapping the question to the correct tables and columns), respectively.

These two failure modes call for two distinct solution mechanisms. Composition fits naturally into a \emph{typed action space}: rather than re-emitting raw SQL at every step, the model operates over a small, closed vocabulary of relational operations whose validity is checked at compile time, eliminating the per-turn SQL-parsing burden and catching column-name errors before execution. Grounding fits naturally into \emph{iteration}: by interacting with the database over multiple rounds---probing value distributions, verifying join keys, sampling rows---the model can disambiguate concepts before committing to a final query. As a secondary benefit, each operation in a typed action space forms a named intermediate, yielding a step-decomposed trace whose reasoning is open to inspection.

To this end, we propose \textbf{ReAct-SQL}, a zero-shot agentic framework \emph{in which the LLM never generates SQL directly}. Given only the database schema and the natural-language question---with no schema-linking module, no candidate prefilling, and no retrieved demonstrations---the model constructs the query through a typed domain-specific language (DSL) over an immutable \emph{Relation} abstraction in a ReAct-style loop \citep{yao2023react}. The DSL targets composition; the loop targets grounding; the resulting trace doubles as a step-decomposed record of the model's reasoning. To evaluate this design comprehensively, we look beyond the end-task accuracy that most prior work emphasizes and additionally measure cost and latency under matched models and identical settings, reporting per-query cost and median latency.
Our contributions are threefold:
\begin{itemize}[leftmargin=*,itemsep=2pt,topsep=4pt]
  \item \textbf{A vanilla ReAct loop suffices for competitive text-to-SQL.} We show that a plain think$\to$act$\to$observe loop---without schema linking, retrieval, candidate sampling, or dedicated verifiers---reaches competitive accuracy at substantially lower cost and latency than more elaborate pipelines.
  \item \textbf{A typed DSL as the LLM's constrained action space.} Targeting the composition problem, we replace free-form SQL generation with a typed DSL: the model is exposed only to 15 schema-validated relational operations (12 Relation-building plus 3 read-only probing) over an immutable \emph{Relation} abstraction, with column references checked at compile time rather than at execution.
  \item \textbf{A comprehensive accuracy--efficiency analysis.} Where prior work focuses primarily on accuracy, we benchmark text-to-SQL pipelines end-to-end under identical conditions and find that recent systems cluster tightly in accuracy yet diverge dramatically in cost and latency, suggesting that future evaluation of deployable text-to-SQL should treat both as first-class metrics alongside accuracy.
\end{itemize}
 
\section{Related Work}
\label{sec:related}

\paragraph{Text-to-SQL Generation.}
Prior systems span three broad approaches. Early neural approaches frame
text-to-SQL as sequence-to-sequence transduction, training encoder--decoder
models on (question, SQL) pairs \citep{zhong2017seq2sql, yu2018spider}. With
the advent of LLMs, in-context learning with schema-aware prompts emerged as
a one-pass alternative \citep{rajkumar2022evaluating}, but
proved brittle on complex, real-world schemas. \emph{Multi-stage pipelines}
address this by decomposing the task into fixed sequential steps, each
targeting a different sub-problem \citep{talaei2024chess, cao2024rsl,
xie2025opensearch,pourreza2023din}. For example, CHESS \citep{talaei2024chess} coordinates
four specialized agents (information retriever, schema selector, candidate
generator, unit tester) in a fixed-order pipeline; RSL-SQL and OpenSearch-SQL
follow a similar pattern with empirically-tuned variations. A more recent
line of work introduces flexible, feedback-driven loops in which the model
dynamically adapts its strategy based on execution results
\citep{nan2026diver, cao2026apex, xie2025sde}. For instance,
DIVER \citep{nan2026diver} performs value linking through interactive probing
tools to ground entity references before SQL generation, and APEX-SQL
\citep{cao2026apex} adds hypothesis--verification loops where the model
proposes candidate SQL, inspects results, and revises over multiple rounds.
In all of these, each turn still emits a complete SQL string. ReAct-SQL
takes the loop further: the model builds the query \emph{incrementally}
inside the loop through a typed DSL, with each step producing a named,
validated intermediate rather than a fresh SQL string.

\paragraph{LLM Agents and ReAct-Style Reasoning.}
The ReAct paradigm interleaves reasoning traces with tool-calling actions \citep{yao2023react} and has since become the dominant design pattern for LLM agents. Subsequent work has extended this loop with self-reflection \citep{shinn2023reflexion}, learned tool use \citep{schick2023toolformer} and persistent skill libraries \citep{wang2023voyager}.

\paragraph{Structured and DSL-Based Generation.}
A parallel line of work imposes structure on LLM outputs to improve reliability. Constrained decoding enforces validity at the token level: PICARD \citep{scholak2021picard} constrains beam search for SQL specifically, and general-purpose frameworks such as Outlines \citep{willard2023outlines} admit regex and context-free grammar constraints. A complementary strategy replaces the output language entirely---CodeAct \citep{wang2024codeact} treats executable Python as the agent action space. Our typed DSL extends this direction: rather than constraining SQL at the token level or reusing a general-purpose language, we define a small, closed vocabulary of 15 relational operations that constitutes the entire action space.

\section{Methodology}
\label{sec:method}

\subsection{Overview}
\label{sec:overview}

\begin{figure*}[t]
  \centering
  \includegraphics[width=1\textwidth]{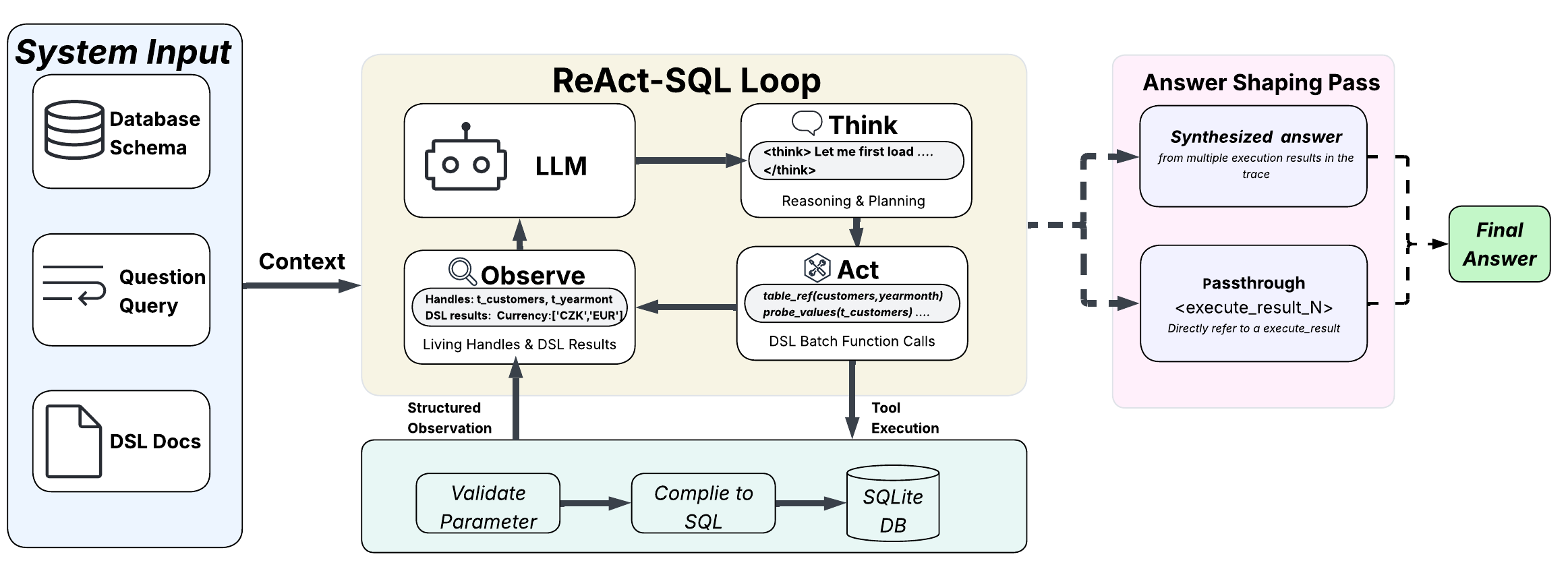}
  \caption{ReAct-SQL architecture. The model receives only the schema, the question, and DSL documentation. It reasons, issues batched DSL operations, and observes execution results in a multi-turn loop. An always-on answer-shaping pass either passes the canonical \texttt{execute} result through (via a \texttt{<execute\_result\_N>} sentinel) or synthesizes a combined multi-metric answer from the trace.}
  \label{fig:architecture}
\end{figure*}

ReAct-SQL takes a deliberately simple approach (Figure~\ref{fig:architecture}): given the database schema and a natural-language question, the model incrementally constructs the query through a ReAct-style \textbf{think$\to$act$\to$observe} loop over a typed relational DSL. At each turn the model first \emph{thinks}, then issues batched DSL operations and observes their execution results, revising until it arrives at an answer. An answer-shaping pass (\S\ref{sec:reconcile}) then selects or synthesizes the canonical answer from the trace.

\subsection{ReAct-Style Query Construction}
The core of ReAct-SQL is a multi-turn \textbf{think$\to$act$\to$observe} loop in which the model constructs the query incrementally. Rather than generating SQL directly---which is difficult to constrain---the model's actions are restricted to a fixed set of typed DSL operations, which a deterministic engine validates and compiles to SQL.

\subsubsection{The Think--Act--Observe Loop}
\label{sec:loop}

At the first turn, the model sees only the natural-language question, the database schema, and the DSL documentation. It first \emph{thinks}: it emits a reasoning block inside \texttt{<think>...</think>} tags that analyzes the question and plans an approach. It then \emph{acts}: it issues a JSON array of DSL tool calls that are executed sequentially, so a handle produced by one call is available to later calls in the same turn. A typical first turn, for instance, loads a few candidate tables with \texttt{table\_ref} and then probes their contents with \texttt{probe\_values} or \texttt{execute}, checking the data against the model's expectations.

Before compilation, the engine validates every parameter against the current schema and the Relation graph (\S\ref{sec:relation}), catching unknown tables, missing columns, and references to columns dropped by earlier operations. On failure it returns a typed error (e.g., \texttt{UnknownColumnError}) with a diagnostic message and skips the rest of the batch; the next \texttt{<think>} block sees which call failed, why, and what succeeded, so the model can reissue a corrected batch. 

For each successful call, the engine returns an \emph{observation}: the output handle name together with its available columns---for example, a \texttt{table\_ref} call returns a handle pointing to the loaded table along with its columns. For \texttt{execute} calls the observation additionally includes a preview of the result rows and the total row count, giving the model concrete data to verify its query logic. Handles persist across turns, so the model can reference any previously created handle later in the trace. The first turn thus has no observation to condition on; from the second turn onward, each \texttt{<think>} block is conditioned on the full environment feedback---the per-call outcome of the previous batch (success or typed error), any \texttt{execute} previews, and every live handle---which the model uses to plan its next batch.

The model itself decides when to stop: once it believes it has reached the answer, it emits a \texttt{final} signal and the loop terminates. We set a configurable turn budget so the model does not waste resources iterating indefinitely on questions it cannot resolve. Because each DSL call is a compact JSON object rather than a full SQL string, the multi-turn loop remains cost-efficient despite its iterative nature.

\subsubsection{Query Construction via Relation Operations}
\label{sec:relation}

Within each turn, the ``act'' phase builds the query by composing operations over an immutable data structure called a \textbf{Relation}. Each Relation represents a SQL subquery together with an \texttt{available\_cols} dictionary mapping logical column names to their underlying SQL expressions. Every DSL operation consumes one or more Relations and returns a \emph{new} Relation rather than mutating its input, so the query-construction history forms a directed acyclic graph (DAG) that the deterministic engine compiles to SQL. Because the model refers to columns by logical name, it is insulated from SQL-level scoping rules---for example, whether a column is visible inside or outside a \texttt{GROUP BY}.

The DSL exposes 12 Relation-building operations and 3 read-only probing operations (Table~\ref{tab:dsl}). The probing operations---\texttt{probe\_values}, \texttt{value\_exists}, and \texttt{sim\_value\_in}, inspired by SDE-SQL's use of SQL probes for self-driven database exploration~\citep{xie2025sde}---let the model sample column values during reasoning to ground entity references and verify value formats before committing to a query. Crucially, each Relation-building operation transforms \texttt{available\_cols} in a well-defined way that mirrors SQL semantics (e.g., \texttt{join\_table} merges both inputs' column dictionaries, while \texttt{aggregate} retains only the group-by and aggregate columns), so at every node in the DAG the engine knows exactly which columns exist. It can therefore validate the model's references \emph{before} generating any SQL---catching typos, stale references from earlier turns, and columns lost through prior projections at compile time rather than at execution.

\begin{table}[!t]
\centering
\footnotesize
\setlength{\tabcolsep}{4pt}
\renewcommand{\arraystretch}{1.0}

\begin{tabular}{@{}lp{4.9cm}@{}}
\toprule
\textbf{Operation} & \textbf{Description} \\
\midrule

\multicolumn{2}{@{}l}{\textit{Relation-building (12)}} \\[-2pt]

\texttt{table\_ref}      & Load a base table as a Relation \\
\texttt{select\_columns} & Project a subset of columns \\
\texttt{distinct}        & Deduplicate rows \\
\texttt{filter\_expr}    & Apply a SQLite WHERE predicate \\
\texttt{join\_table}     & Inner / left / cross join \\
\texttt{aggregate}       & \texttt{GROUP BY} with aggregations \\
\texttt{derive\_column}  & Add a column from an SQLite expression \\
\texttt{window\_column}  & Add a rank/aggregate column; keeps all rows \\
\texttt{order\_by}       & Sort by one or more keys \\
\texttt{limit}           & Truncate to $n$ rows \\
\texttt{materialize}     & Persist to \texttt{TEMP TABLE}; reset depth \\
\texttt{execute}         & Compile and run; return rows \\

\midrule

\multicolumn{2}{@{}l}{\textit{Probing, read-only (3)}} \\[-2pt]

\texttt{probe\_values}   & Sample distinct values from a column \\
\texttt{value\_exists}   & Test whether a value exists \\
\texttt{sim\_value\_in}  & Fuzzy-match a value against a column \\

\bottomrule
\end{tabular}
\caption{The 12 relation-building and 3 read-only probing operations of the DSL.}
\label{tab:dsl}

\vspace{-3mm}

\end{table}

\subsubsection{Answer Shaping}
\label{sec:reconcile}

Although the iteration loop already produces an answer, a lightweight
\emph{answer-shaping} pass presents it unambiguously, without further
reasoning. From the question, the reasoning trace, and previews of prior
\texttt{execute} results, the shaper emits either a synthesized table---when
the answer spans several executes, common for BIRD's multi-part
questions---or a single-line sentinel \texttt{<execute\_result\_N>} that
designates the \mbox{$N$-th} execute as the answer (e.g., rewriting an
implicitly returned row list as an explicit yes/no). The sentinel keeps
shaping cost independent of result size and preserves the engine's output
\emph{byte-for-byte}, avoiding the surface-level mismatches that motivate
our LLM-judge scoring (\S\ref{sec:judge}).

\section{Experimental Setup}
\label{sec:experiments}

\subsection{Datasets}
\label{sec:dataset}

\paragraph{BIRD mini-dev (corrected).}
We evaluate on the corrected BIRD mini-dev \citep{jin2026pervasive}, a 498-question subset of BIRD \citep{li2023bird} spanning 11 databases with per-question difficulty labels. A recent audit finds that a substantial fraction of the original mini-dev questions contain annotation errors (hint--SQL contradictions, missing filters, ambiguous targets), and that correcting them can shift per-system accuracy markedly between the two releases, so conclusions about system ordering on BIRD depend substantially on which release one uses.

\paragraph{EHR-SQL.}
To test whether conclusions on a general-purpose benchmark survive a shift in schema domain, we additionally evaluate on EHR-SQL \citep{lee2022ehrsql}, a clinical text-to-SQL benchmark, using the MIMIC-IV port released for the EHRSQL 2024 shared task \citep{lee2024ehrsql} (934 answerable test questions; the unanswerable subset is excluded). Its ${\sim}$15 interconnected tables (e.g., \texttt{d\_icd\_diagnoses}, \texttt{inputevents}, \texttt{labevents}, \texttt{chartevents}) follow medical conventions that differ sharply from BIRD's general-purpose schemas. We deliberately withhold EHR-SQL's labeled training split from all systems, so no question--SQL pairs are available for in-context retrieval---a setting that mimics industrial deployment on a specialized database and that especially stresses retrieval-augmented pipelines whose retrievers are trained on BIRD-style data.

\subsection{LLM-Judge Scoring}
\label{sec:judge}

The official BIRD protocol scores a prediction as correct only if \texttt{set(cursor.fetchall())} on the predicted SQL exactly matches the gold result---a criterion that penalizes numerical drift, extra harmless projection columns, and trivial row-ordering differences, systematically understating the accuracy of agentic systems whose outputs are semantically correct but not byte-identical \citep{huo2026bird, zheng2023judging}. EHR-SQL exhibits analogous mismatches in date encodings, value casing, and identifier prefixes. We replace strict equality with a fixed LLM-as-judge scorer---applied uniformly to every system on both benchmarks, returning a binary \texttt{MATCH}/\texttt{MISMATCH} verdict and instructed to treat such surface differences as \texttt{MATCH}. Because a single judge is used across all systems, any residual bias affects them symmetrically. The detailed implementation is in Appendix~\ref{app:judge_prompt}.

\subsection{Baselines}
\label{sec:baselines}

We compare against five text-to-SQL systems spanning the design spectrum: \textbf{CHESS} \citep{talaei2024chess}, \textbf{RSL-SQL} \citep{cao2024rsl}, \textbf{DSR-SQL} \citep{huang2025dsr}, \textbf{DIN-SQL} \citep{pourreza2023din}, and \textbf{MAC-SQL} \citep{wang2024macsql}; full descriptions are in Appendix~\ref{app:baselines}. RSL-SQL is the only baseline that uses retrieved in-context examples---via a sentence-transformer retriever trained on BIRD question--SQL pairs---and we keep its default retrieval configuration since it is core to the method. All systems use Gemini~2.5~Flash \citep{geminiteam2025gemini} as backbone. To assess the generalizability of ReAct-SQL, we additionally evaluate it across a range of backbones---from Gemini-2.5-Flash-Lite to Gemini-2.5-Pro\citep{geminiteam2025gemini} and a non-Gemini family (Claude Sonnet 4.6\citep{anthropic2026claude46}).

\section{Results}
\label{sec:results}

\subsection{Main Results}
\label{sec:main_results}

Table~\ref{tab:main_results} reports the controlled comparison on both benchmarks.
\begin{table*}[t]
\centering
\small
\begin{tabular}{@{}lcccccc@{}}
\toprule
& \multicolumn{3}{c}{\textbf{BIRD mini-dev (498q)}} & \multicolumn{3}{c}{\textbf{EHR-SQL (934q)}} \\
\cmidrule(lr){2-4}\cmidrule(lr){5-7}
\textbf{System} & \textbf{Acc.\ (\%)} & \textbf{Cost (\$/q)} & \textbf{Med.\ Lat.\ (s)} & \textbf{Acc.\ (\%)} & \textbf{Cost (\$/q)} & \textbf{Med.\ Lat.\ (s)} \\
\midrule
CHESS \citep{talaei2024chess}              & \textbf{85.3} & 0.0842 & 53.7 & 70.1 & 0.1140 & 104.8 \\
RSL-SQL$^{\dagger}$ \citep{cao2024rsl}     & 84.1 & \textbf{0.0053} & 22.4 & 64.8 & \textbf{0.0087} & 40.2 \\
DSR-SQL \citep{huang2025dsr}                 & 84.9 & 0.0252 & 28.1 & 69.2 & 0.0304 & 38.8 \\
DIN-SQL \citep{pourreza2023din}            & 70.3 & 0.0152 & 13.6 & 54.0 & 0.0193 & 17.7 \\
MAC-SQL \citep{wang2024macsql}             & 66.1 & 0.0026 & 13.0 & 41.0 & 0.0026 & 18.8 \\
\midrule
ReAct-SQL (ours)                           & 84.5 & 0.0107 & \textbf{7.5} & \textbf{73.9} & 0.0166 & \textbf{13.2} \\
\bottomrule
\end{tabular}
\caption{Main results on corrected BIRD mini-dev \citep{jin2026pervasive} and EHR-SQL \citep{lee2022ehrsql}; all systems receive identical hints and are scored by the same LLM judge. Latency is the per-query median; query latency is right-skewed, so the median is the representative measure---full statistics (mean, min, max, and coefficient of variation) are in Appendix~\ref{app:latency}. Cost is the per-query mean: it depends only on input and output token counts, unlike latency, which also reflects server load and network conditions. $^{\dagger}$Uses retrieved in-context demonstrations from BIRD training pairs; all other systems are zero-shot.}
\label{tab:main_results}
\end{table*}
On the corrected BIRD mini-dev (Table~\ref{tab:main_results}, left half), four of the six systems land within 1.2 accuracy points of one another---CHESS (85.3\%), DSR-SQL (84.9\%), ReAct-SQL (84.5\%), and RSL-SQL (84.1\%)---while DIN-SQL and MAC-SQL trail 14--18 points behind. Where accuracy no longer separates systems, efficiency becomes the discriminating axis, and ReAct-SQL is the fastest of all six: a 7.5\,s median latency, \textbf{7.2$\times$ faster than CHESS} and \textbf{3.8$\times$ faster than DSR-SQL}, the two systems it ties on accuracy. Both reach that accuracy through elaborate machinery---CHESS through extensive candidate sampling (3 candidates $\times$ 5 unit tests over separate retrieval and unit-testing stages, ${\sim}$20 LLM calls per question), DSR-SQL through a two-state pipeline with an explicit schema-refinement phase. ReAct-SQL matches them with a single ReAct loop and none of that scaffolding. It is not the cheapest, however: RSL-SQL reaches comparable accuracy at roughly half the per-query cost (\$0.0053 vs.\ \$0.0107), leveraging in-context examples retrieved from the BIRD training set.

On EHR-SQL (Table~\ref{tab:main_results}, right half), the BIRD parity breaks in ReAct-SQL's favor. ReAct-SQL leads accuracy outright at \textbf{73.9\%}---3.8 points above CHESS (70.1\%) and 4.7 above DSR-SQL (69.2\%)---while remaining the fastest system, at a 13.2\,s median. The clinical schema costs every system accuracy, but ReAct-SQL pays the least: its 10.6-point drop from BIRD is the smallest of the six, against 15--25 points for the rest (CHESS $-15.2$, DSR-SQL $-15.7$, DIN-SQL $-16.3$, RSL-SQL $-19.3$, MAC-SQL $-25.1$). The elaborate pipelines that matched ReAct-SQL on BIRD do not carry that parity to the harder schema; their extra machinery does not translate into robustness. RSL-SQL shows this most starkly: its BIRD-competitive accuracy leans on in-context examples retrieved from BIRD training pairs, and on the clinical schema those examples are out-of-domain---its accuracy falls 19.3 points, the steepest drop among the four systems competitive on BIRD. A pipeline that depends on retrieved demonstrations is brittle when applied to a new database for which no labeled examples exist; ReAct-SQL's single iterate-and-probe loop, fully zero-shot, both generalizes better and runs nearly $8\times$ faster than CHESS.

\begin{figure*}[t]
  \centering
  \includegraphics[width=0.75\textwidth]{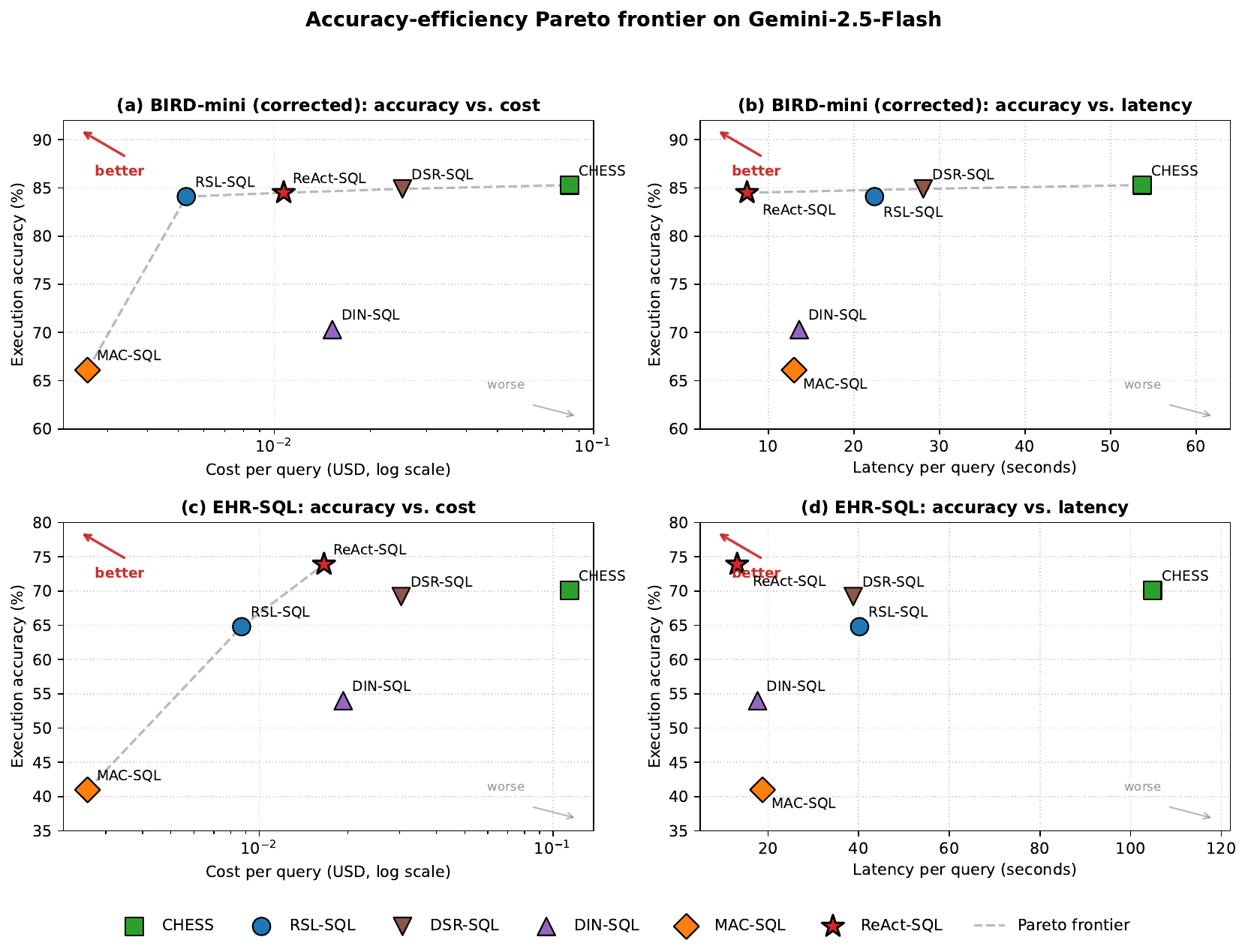}
  \caption{Accuracy-efficiency Pareto frontier on Gemini-2.5-Flash. Each point is one system on one benchmark. Upper-left is better (higher accuracy, lower cost/latency).}
  \label{fig:pareto}
\end{figure*}

Figure~\ref{fig:pareto} plots all six systems against cost and latency on both benchmarks; ReAct-SQL lies on the Pareto frontier in all four panels---no other system is simultaneously cheaper (or faster) and at least as accurate. RSL-SQL undercuts it on BIRD cost, but only at lower accuracy. Altogether, ReAct-SQL remains competitive on both axes across the two benchmarks.

\paragraph{Cross-backbone consistency.}
To confirm the framework generalizes beyond Gemini-2.5-Flash, we additionally evaluate ReAct-SQL on a 200-question subset of EHR-SQL across four backbones spanning two model families (Gemini-2.5-Flash-Lite, Flash, Pro, and Claude Sonnet 4.6). ReAct-SQL improves over the zero-shot raw-SQL baseline at every scale, with gains ranging from $+15.0$ to $+36.5$ points (Table~\ref{tab:backbone_generalization}); per-backbone analysis is in Appendix~\ref{app:backbone}.

\begin{table}[t]
\centering
\footnotesize
\setlength{\tabcolsep}{4pt}
\begin{tabular}{@{}lrrr@{}}
\toprule
\textbf{Backbone} & \textbf{Base.} & \textbf{+ReAct-SQL} & \textbf{Gain} \\
\midrule
Gemini-2.5-Flash-Lite & 30.5 & 45.5 & $+15.0$ \\
Gemini-2.5-Flash      & 44.0 & 77.0 & $+33.0$ \\
Gemini-2.5-Pro        & 45.0 & \textbf{81.5} & $\mathbf{+36.5}$ \\
Claude Sonnet 4.6     & \textbf{53.0} & \textbf{87.0} & $+34.0$ \\
\bottomrule
\end{tabular}
\caption{ReAct-SQL vs.\ the zero-shot raw-SQL baseline on a 200-question subset of EHR-SQL, across four backbones from two model families. Accuracy in \%. As a subset, absolute accuracies differ slightly from the full-benchmark run.}
\label{tab:backbone_generalization}
\end{table}

\section{Ablation Study}
\label{sec:ablation}

ReAct-SQL combines two design choices: \emph{iteration} (the ReAct loop
with execution feedback) and the \emph{typed DSL} (a closed action space
with compile-time validation). We disentangle their contributions with a
three-setting incremental ablation on both benchmarks (corrected BIRD
mini-dev and EHR-SQL), under Gemini-2.5-Flash
and the same LLM-judge scorer used in \S\ref{sec:results}. We additionally report McNemar's paired tests and paired bootstrap $95\%$ confidence intervals on each pairwise accuracy difference; the full table of pairwise comparisons is in Appendix~\ref{app:stats}.

\begin{itemize}[leftmargin=*,itemsep=2pt,topsep=4pt]
  \item \textbf{Baseline (zero-shot raw SQL).} A plain one-shot baseline:
        schema + question in, SQL out, single call.
  \item \textbf{+ Iteration (raw-SQL ReAct).} The ReAct loop runs as in
        the full system, but each ``act'' is a raw \texttt{execute\_sql(...)}
        call rather than a typed DSL operation.
  \item \textbf{+ DSL (full ReAct-SQL).} The complete system of
        \S\ref{sec:method}. Adds the typed DSL on top of iteration.
\end{itemize}

\begin{table}[t]
\centering
\footnotesize
\setlength{\tabcolsep}{4pt}
\begin{tabular}{lcc}
\toprule
\textbf{Setting} & \textbf{BIRD} & \textbf{EHR-SQL} \\
& ($n{=}498$) & ($n{=}934$) \\
\midrule
Baseline (zero-shot raw SQL) & 78.5 & 45.9 \\
+ Iteration (raw-SQL ReAct) & 79.1 & 72.2 \\
+ Iteration + DSL (ReAct-SQL) & \textbf{84.5} & \textbf{73.9} \\
\bottomrule
\end{tabular}
\caption{Incremental ablation on Gemini-2.5-Flash; accuracy in \%, additional analysis of McNemar tests and bootstrap CIs are listed in Appendix~\ref{app:stats}}
\label{tab:ablation}
\end{table}
\paragraph{Result overview.}
The ablation in Table~\ref{tab:ablation} reveals two clear patterns. The
two benchmarks start from very different baselines: zero-shot raw SQL
reaches $78.5\%$ on BIRD but only $45.9\%$ on EHR-SQL, a $33$-point gap
that foreshadows their distinct difficulty profiles. On EHR-SQL, adding iteration alone lifts accuracy from $45.9\%$ to $72.2\%$, a $+26.2$-point gain that is highly significant (McNemar $p < 10^{-45}$); adding the DSL on top yields only a marginal $+1.7$ points. On BIRD, the pattern inverts: iteration alone moves the needle by only $+0.6$ points, while adding the DSL on top contributes $+5.4$ points (McNemar $p = 0.006$).

\paragraph{Two bottlenecks, two remedies.}
On \emph{EHR-SQL}, the bottleneck is value-level uncertainty: MIMIC-IV's
clinical conventions (lower-case drug names like \texttt{metformin},
simulated year-$2100$ timestamps, abbreviated table names
\texttt{d\_icd\_diagnoses}/\texttt{hadm\_id}) lie far from a model's
pretraining distribution, so a zero-shot model writes SQL ``blind'' and
has no way to recover from a wrong-cased value or wrong table.
Iteration breaks this open: the model probes the database
(\texttt{probe\_values}, \texttt{value\_exists}, \texttt{sim\_value\_in}),
observes what is stored, and only then writes the query; once probing
has grounded the values, raw SQL and typed DSL express the same logic
equally well, which is why the DSL adds little on top. On \emph{BIRD},
the bottleneck is structural correctness: its general-purpose schemas
are well-represented in pretraining and value uncertainty is small, but
many questions require parallel sub-computations across segments---e.g.,
``percentage increase in consumption for SME, LAM, and KAM
respectively''---that compile to deeply nested SQL. Raw-SQL ReAct
cannot easily recover from a structural mistake in one
execution-feedback round (a botched join forces a from-scratch
rewrite); the typed DSL targets this directly, abstracting SQL grammar
away and letting the model name the logic step by step in a closed
vocabulary, with compile-time validation and the \texttt{materialize}
primitive preventing the structural errors raw-SQL ReAct incurs. 

In summary, iteration and the typed DSL are each indispensable to ReAct-SQL: iteration addresses the grounding bottleneck (dominant on EHR-SQL) and the DSL the composition bottleneck (dominant on BIRD). Their relative impact thus shifts with the dataset, but only combining the two delivers the best performance on both benchmarks.

\section{Analysis and Implications}
\label{sec:discussion}

\paragraph{Iteration subsumes the grounding pipeline.}
Prior text-to-SQL systems devote substantial machinery to a single goal: giving the model enough knowledge of the database before it writes SQL. Schema-linking modules prune the schema; retrievers fetch similar question--SQL pairs; dedicated exploration or schema-refinement stages surface values and conventions. Each is a static preprocessing step whose output is fixed before generation begins. We show that this class of machinery can instead be folded into the model itself, by casting text-to-SQL in the LLM-agent paradigm: the model is handed tools to inspect its environment---the database---and left to resolve uncertainty on its own. On EHR-SQL this alone---the ReAct loop, with no schema linker, retrieval, or exploration module---lifts accuracy from 45.9\% to 72.2\% ($+26.2$ points), with far less engineering overhead than a dedicated preprocessing stack. The reason is that grounding is fundamentally a question the model should ask on demand, not one answered in advance. Real databases carry latent conventions---abbreviations, categorical encodings, timestamp formats, institution-specific semantics---that no amount of static schema text reliably conveys. Inside the loop, the model probes the live database exactly when it is uncertain, observes what is stored, and only then commits---and this behavior emerges from a general ReAct loop rather than an engineered stage.

\paragraph{A constrained action space yields reliable SQL generation.}
SQL queries can grow unboundedly complex---nested subqueries, multi-way joins, layered aggregation---yet that complexity is assembled from a small, recurring set of relational operations. The typed DSL exposes exactly that closed set: the model manages the \emph{logic} of a query---load a table, join, filter, aggregate---and never emits raw SQL grammar. This matters most where free-form generation is brittle: structurally complex queries. The standard remedy there---generate a full query, execute it, feed back the error, regenerate---rewrites the entire query on every failure, so a fixed mistake can introduce a fresh one and recovery is unreliable. A constrained action space changes the failure mode: because the model builds the query one typed operation at a time rather than emitting it whole, a hallucination---a reference to a non-existent table, say---is confined to the single operation that contains it, and the model corrects just that step. Composition becomes a sequence of localized, individually-valid steps---an implicit chain-of-thought over executable relational states rather than SQL tokens.

\paragraph{Why this matters: accuracy has saturated, efficiency has not.}
Recent text-to-SQL systems have steadily improved accuracy by adding components---retrieval, candidate sampling, verification, multi-agent orchestration---and these advances are real. But on corrected BIRD mini-dev, accuracy has largely converged: four systems now sit within 1.2 points of one another, while their latency spans nearly an order of magnitude. At this point accuracy no longer separates strong systems; efficiency does. Each added component carries an inherent cost---candidate sampling multiplies LLM calls, multi-agent orchestration multiplies prompts, retrieval and verification add stages---and that cost is now the main axis of variation among comparable systems. ReAct-SQL is a data point in the other direction: it reaches the same accuracy band with a single agentic loop, at a fraction of the cost and latency. We do not argue that elaborate components are unnecessary---many bring genuine accuracy gains---only that their efficiency cost deserves to be treated as a first-class deployment consideration, and that achieving the benefits of such components at low cost and latency is an important open problem.

\section{Conclusion}
\label{sec:conclusion}

We have presented \textbf{ReAct-SQL}, a zero-shot text-to-SQL framework
with a deliberately minimal design---just a typed DSL
with compile-time validation and a ReAct loop with execution feedback.
ReAct-SQL reaches competitive accuracy on corrected BIRD mini-dev and
EHR-SQL while running up to $8\times$ faster than the most elaborate
baselines. An incremental ablation
isolates two orthogonal bottlenecks---\emph{grounding}, solved by
iteration, and \emph{composition}, solved by the DSL. Our experiments further suggest two broader takeaways: the DSL implicitly decomposes reasoning into step-by-step relational operations rather than a monolithic SQL string, and the ReAct loop naturally subsumes specialized preprocessing stages (schema linking, value exploration) that existing pipelines handle as separate modules. Our
cross-system efficiency analysis surfaces a methodological point that
current research often overlooks: pipelines within a few accuracy
points of one another can differ by more than an order of magnitude in
cost and latency (Figure~\ref{fig:pareto}). We hope to encourage future
text-to-SQL research to evaluate not on accuracy alone, but on the full
accuracy--cost--latency trade-off.

\section*{Limitations}
Despite the promising results of ReAct-SQL, our approach has limitations
worth clarifying. First, like any agentic pipeline, ReAct-SQL
relies heavily on the base model's ability to follow structured
instructions; weaker models tend to hallucinate non-existent operations
or emit malformed DSL calls, which substantially degrades performance.
Future work will explore specialized fine-tuning (using the traces
collected in this work) to produce a lightweight model proficient at
DSL-based query construction. Second, our evaluation does not
cover all dimensions of real-world deployment. The schemas we test are
moderately sized, where the ReAct loop's interactive exploration
successfully subsumes explicit schema linking; for enterprise databases
with thousands of tables, explicit schema pruning before the loop
remains necessary simply to fit the schema within the model's context
window. Extending ReAct-SQL to enterprise-scale schemas and
conversational deployment is a natural follow-up. Third, the
formal expressiveness of the DSL is an open question. In our
experiments the DSL is empirically sufficient---we observed no failure
attributable to inexpressibility---but a rigorous mathematical proof
that the DSL is complete over standard relational operations remains
future work.

\section*{Ethical considerations}

\paragraph{Use of artifacts.}
This work uses only publicly available research artifacts, and our use of
each is consistent with its intended research purpose. The BIRD benchmark and
its corrected mini-dev split, together with the EHR-SQL benchmark, are
publicly released for research use, and we use them solely to evaluate
text-to-SQL systems---the task for which they were designed. The baseline
systems we compare against---CHESS, RSL-SQL, DSR-SQL, DIN-SQL, and
MAC-SQL---are run from their authors' public code releases under their
respective open-source licenses. We will release ReAct-SQL as open-source
code for research use.

\paragraph{Clinical data.}
Our EHR-SQL experiments use the MIMIC-IV port released for the EHRSQL~2024
shared task. MIMIC-IV is distributed under the PhysioNet Credentialed Health
Data License and requires credentialed access, completion of human-subjects
research training, and acceptance of a data use agreement. The database is
fully de-identified; we use it strictly for the non-commercial research
purpose for which it is intended and make no attempt to re-identify
individuals. No new human data was collected for this work.

\bibliography{custom}

@inproceedings{yu2018spider,
  title = {Spider: A Large-Scale Human-Labeled Dataset for Complex and Cross-Domain Semantic Parsing and Text-to-SQL Task},
  author = {Yu, Tao and Zhang, Rui and Yang, Kai-Chou and Yasunaga, Michihiro and Wang, Dongxu and Li, Zifan and Ma, James and Li, Irene Z and Yao, Qingning and Roman, Shanelle and others},
  year = {2018},
  journal = {Conference on Empirical Methods in Natural Language Processing},
  pages = {3911-3921},
  doi = {10.18653/v1/D18-1425},
  publisher = {Association for Computational Linguistics},
}

@inproceedings{li2023bird,
  author = {Li, Jinyang and Hui, Binyuan and Qu, Ge and Li, Binhua and Yang, Jiaxi and Li, Bowen and Wang, Bailin and Qin, Bowen and Cao, Rongyu and Geng, Ruiying and others},
  title = {Can LLM Already Serve as A Database Interface? A Big Bench for Large-Scale Database Grounded Text-to-SQLs},
  booktitle = {Advances in Neural Information Processing Systems},
  year = {2023},
  pages = {42330--42357}
}

@inproceedings{huo2026bird,
  title = {BIRD-INTERACT: Re-imagining Text-to-SQL Evaluation for Large Language Models via Lens of Dynamic Interactions},
  author = {Huo, Nan and Xu, Xiaohan and Li, Jinyang and Jacobsson, Per and Lin, Shi and Qin, Bowen and Hui, Binyuan and Li, Xiaolong and Qu, Ge and Si, Shuzheng and others},
  year = {2025},
  journal = {arXiv.org},
  doi = {10.48550/arXiv.2510.05318},
}

@inproceedings{zhong2017seq2sql,
  title = {Seq2SQL: Generating Structured Queries from Natural Language using Reinforcement Learning},
  author = {Zhong, Victor and Xiong, Caiming and Socher, Richard},
  booktitle = {arXiv.org},
  year = {2017},
  journal = {arXiv.org},
}

@article{rajkumar2022evaluating,
  title = {Evaluating the Text-to-SQL Capabilities of Large Language Models},
  author = {Rajkumar, Nitarshan and Li, Raymond and Bahdanau, Dzmitry},
  journal = {arXiv.org},
  year = {2022},
  doi = {10.48550/arXiv.2204.00498},
}

@inproceedings{pourreza2023din,
  title = {DIN-SQL: Decomposed In-Context Learning of Text-to-SQL with Self-Correction},
  author = {Pourreza, Mohammadreza and Rafiei, Davood},
  booktitle = {Neural Information Processing Systems},
  year = {2023},
  journal = {Neural Information Processing Systems},
  pages = {36339-36348},
  doi = {10.48550/arXiv.2304.11015},
  publisher = {Neural Information Processing Systems Foundation, Inc. (NeurIPS)},
}

@article{wang2024macsql,
  title = {MAC-SQL: A Multi-Agent Collaborative Framework for Text-to-SQL},
  author = {Bing Wang and Changyu Ren and Jian Yang and Xinnian Liang and Jiaqi Bai and Qian-Wen Zhang and Zhao Yan and Zhoujun Li},
  year = {2023},
  journal = {International Conference on Computational Linguistics},
  doi = {10.48550/arXiv.2312.11242},
}

@inproceedings{yao2023react,
  title = {ReAct: Synergizing Reasoning and Acting in Language Models},
  author = {Yao, Shunyu and Zhao, Jeffrey and Yu, Dian and Du, Nan and Shafran, Izhak and Narasimhan, Karthik and Cao, Yuan},
  booktitle = {International Conference on Learning Representations (ICLR)},
  year = {2023}
}

@article{cao2026apex,
  title = {APEX-SQL: Talking to the data via Agentic Exploration for Text-to-SQL},
  author = {Cao, Bowen and Liao, Weibin and Sun, Yushi and Fang, Dong and Li, Haitao and Lam, Wai},
  journal = {arXiv.org},
  year = {2026},
  doi = {10.48550/arXiv.2602.16720},
}

@article{xie2025opensearch,
  title = {OpenSearch-SQL: Enhancing Text-to-SQL with Dynamic Few-Shot and Consistency Alignment},
  author = {Xie, Xiangjin and Xu, Guangwei and Zhao, Lingyan and Guo, Ruijie},
  journal = {Proc. ACM Manag. Data},
  year = {2025},
  volume = {3},
  number = {3},
  pages = {1-24},
  doi = {10.1145/3725331},
  publisher = {Association for Computing Machinery (ACM)},
}

@article{huang2025dsr,
  title = {Text-to-SQL as Dual-State Reasoning: Integrating Adaptive Context and Progressive Generation},
  author = {Hao, Zhifeng and Song, Qibin and Cai, Ruichu and Xu, Boyan},
  journal = {arXiv.org},
  year = {2025},
  doi = {10.48550/arXiv.2511.21402},
}

@article{talaei2024chess,
  title = {CHESS: Contextual Harnessing for Efficient SQL Synthesis},
  author = {Talaei, Shayan and Pourreza, Mohammadreza and Chang, Yu-Chen and Mirhoseini, Azalia and Saberi, Amin},
  journal = {arXiv.org},
  year = {2024},
  doi = {10.48550/arXiv.2405.16755},
}

@article{cao2024rsl,
  title = {RSL-SQL: Robust Schema Linking in Text-to-SQL Generation},
  author = {Zhenbiao Cao and Yuanlei Zheng and Zhihao Fan and Xiaojin Zhang and Wei Chen and Xiang Bai},
  year = {2024},
  journal = {arXiv.org},
  doi = {10.48550/arXiv.2411.00073},
}

@article{nan2026diver,
  title = {DIVER: A Robust Text-to-SQL System with Dynamic Interactive Value Linking and Evidence Reasoning},
  author = {Nan, Yafeng and Sun, Haifeng and Zhuang, Zirui and Qi, Qi and Chu, Guojun and Liao, Jianxin and Pei, Dan and Wang, Jingyu},
  journal = {Proceedings of the ACM on Management of Data},
  year = {2026},
  volume = {4},
  number = {1},
  pages = {1-24},
  doi = {10.1145/3786640},
  publisher = {Association for Computing Machinery (ACM)},
}

@article{xie2025sde,
  title = {SDE-SQL: Enhancing Text-to-SQL Generation in Large Language Models via Self-Driven Exploration with SQL Probes},
  author = {Xie, Wenxuan and Dai, Yaxun and Jiang, Wenhao},
  journal = {arXiv.org},
  year = {2025},
  doi = {10.48550/arXiv.2506.07245},
}

@inproceedings{jin2026pervasive,
  title = {Pervasive Annotation Errors Break Text-to-{SQL} Benchmarks and Leaderboards},
  author = {Jin, Tengjun and Choi, Yoojin and Zhu, Yuxuan and Kang, Daniel},
  booktitle = {Proceedings of the VLDB Endowment},
  year = {2026},
  note = {arXiv:2601.08778. Corrected BIRD mini-dev: \url{https://github.com/uiuc-kang-lab/text_to_sql_benchmarks}},
  eprint = {2601.08778},
  archiveprefix = {arXiv},
  primaryclass = {cs.AI},
  journal = {Proceedings of the VLDB Endowment},
  volume = {19},
  number = {5},
  pages = {931-944},
  doi = {10.48550/arXiv.2601.08778},
  publisher = {Association for Computing Machinery (ACM)},
}

@article{zheng2023judging,
  title = {Judging {LLM}-as-a-Judge with {MT}-Bench and Chatbot Arena},
  author = {Zheng, Lianmin and Chiang, Wei-Lin and Sheng, Ying and Zhuang, Siyuan and Wu, Zhanghao and Zhuang, Yonghao and Lin, Zi and Li, Zhuohan and Li, Dacheng and Xing, Eric P. and others},
  journal = {Neural Information Processing Systems},
  year = {2023},
  pages = {46595-46623},
  doi = {10.52202/075280-2020},
  publisher = {Neural Information Processing Systems Foundation, Inc. (NeurIPS)},
}

@inproceedings{shinn2023reflexion,
  title = {Reflexion: Language Agents with Verbal Reinforcement Learning},
  author = {Shinn, Noah and Cassano, Federico and Labash, Beck and Gopinath, A. and Narasimhan, Karthik and Yao, Shunyu},
  booktitle = {Neural Information Processing Systems},
  year = {2023},
  journal = {Neural Information Processing Systems},
  pages = {8634-8652},
  doi = {10.52202/075280-0377},
  publisher = {Neural Information Processing Systems Foundation, Inc. (NeurIPS)},
}

@inproceedings{schick2023toolformer,
  title = {Toolformer: Language Models Can Teach Themselves to Use Tools},
  author = {Schick, Timo and Dwivedi-Yu, Jane and Dessì, Roberto and Raileanu, R. and Lomeli, M. and Zettlemoyer, Luke and Cancedda, Nicola and Scialom, Thomas},
  booktitle = {Neural Information Processing Systems},
  year = {2023},
  journal = {Neural Information Processing Systems},
  pages = {68539-68551},
  doi = {10.48550/arXiv.2302.04761},
  publisher = {Neural Information Processing Systems Foundation, Inc. (NeurIPS)},
}

@article{wang2023voyager,
  title = {Voyager: An Open-Ended Embodied Agent with Large Language Models},
  author = {Wang, Guanzhi and Xie, Yuqi and Jiang, Yunfan and Mandlekar, Ajay and Xiao, Chaowei and Zhu, Yuke and Fan, Linxi and Anandkumar, Anima},
  journal = {Trans. Mach. Learn. Res.},
  year = {2023},
  doi = {10.48550/arXiv.2305.16291},
}

@inproceedings{wang2024codeact,
  title = {Executable Code Actions Elicit Better {LLM} Agents},
  author = {Wang, Xingyao and Chen, Yangyi and Yuan, Lifan and Zhang, Yizhe and Li, Yunzhu and Peng, Hao and Ji, Heng},
  booktitle = {International Conference on Machine Learning},
  year = {2024},
  journal = {International Conference on Machine Learning},
  doi = {10.48550/arXiv.2402.01030},
}

@inproceedings{scholak2021picard,
  title = {{PICARD}: Parsing Incrementally for Constrained Auto-Regressive Decoding from Language Models},
  author = {Scholak, Torsten and Schucher, Nathan and Bahdanau, Dzmitry},
  booktitle = {Conference on Empirical Methods in Natural Language Processing},
  year = {2021},
  journal = {Conference on Empirical Methods in Natural Language Processing},
  pages = {9895-9901},
  doi = {10.18653/v1/2021.emnlp-main.779},
  publisher = {Association for Computational Linguistics},
}

@article{willard2023outlines,
  title = {Efficient Guided Generation for Large Language Models},
  author = {Willard, Brandon T. and Louf, R{\'e}mi},
  journal = {arXiv.org},
  year = {2023},
  doi = {10.48550/arXiv.2307.09702},
}

@article{geminiteam2025gemini,
  title = {{Gemini 2.5}: Pushing the Frontier with Advanced Reasoning, Multimodality, Long Context, and Next Generation Agentic Capabilities},
  author = {Comanici, Gheorghe and Bieber, E. and Schaekermann, Mike and Pasupat, Ice and Sachdeva, Noveen and Dhillon, Inderjit S. and Blistein, Marcel and Ram, Ori and Zhang, Dan and Rosen, Evan and others},
  journal = {arXiv.org},
  year = {2025},
}

@inproceedings{lee2022ehrsql,
  title = {EHRSQL: A Practical Text-to-SQL Benchmark for Electronic Health Records},
  author = {Lee, Gyubok and Hwang, Hyeonji and Bae, Seongsu and Kwon, Yeonsu and Shin, Woncheol and Yang, Seongjun and Seo, Minjoon and Kim, Jong-Yeup and Choi, Edward},
  booktitle = {Neural Information Processing Systems},
  year = {2023},
  journal = {Neural Information Processing Systems},
  doi = {10.48550/arXiv.2301.07695},
}

@inproceedings{lee2024ehrsql,
  title = {Overview of the EHRSQL 2024 Shared Task on Reliable Text-to-SQL Modeling on Electronic Health Records},
  author = {Lee, Gyubok and Kweon, Sunjun and Bae, Seongsu and Choi, Edward},
  booktitle = {Clinical Natural Language Processing Workshop},
  year = {2024},
  eprint = {2405.06673},
  archiveprefix = {arXiv},
  primaryclass = {cs.CL},
  journal = {Clinical Natural Language Processing Workshop},
  pages = {644-654},
  doi = {10.48550/arXiv.2405.06673},
  publisher = {Association for Computational Linguistics},
}

@misc{anthropic2026claude46,
  title        = {Claude Sonnet 4.6 System Card},
  author       = {{Anthropic}},
  year         = {2026},
  howpublished = {\url{https://www-cdn.anthropic.com/78073f739564e986ff3e28522761a7a0b4484f84.pdf}},
  note         = {Accessed: 2026-05-26}
}
\appendix
\section{Statistical Analysis}
\label{app:stats}

\begin{table*}[h]
\centering
\small
\begin{tabular}{@{}lrrrrl@{}}
\toprule
\textbf{Comparison} & \textbf{$\Delta$Acc} & \textbf{95\% CI} & \textbf{$b$} & \textbf{$c$} & \textbf{$p$} \\
\midrule
\multicolumn{6}{@{}l}{\textit{BIRD mini-dev (n = 498)}} \\
Baseline $\to$ +Iter   & $+0.60$  & $[-3.41,+4.62]$ & 53 & 56 & $0.848$ (n.s.) \\
+Iter $\to$ +Iter+DSL       & $+5.42$  & $[+1.61,+9.24]$ & 32 & 59 & $0.006$ \\
Baseline $\to$ +Iter+DSL    & $+6.02$  & $[+2.61,+9.64]$ & 28 & 58 & $0.002$ \\
\midrule
\multicolumn{6}{@{}l}{\textit{EHR-SQL (n = 934)}} \\
Baseline $\to$ +Iter   & $+26.23$ & $[+22.91,+29.66]$ &  38 & 283 & ${\sim}10^{-47}$ \\
+Iter $\to$ +Iter+DSL       & $+1.71$  & $[-1.07,+4.50]$   &  78 &  94 & $0.253$ (n.s.) \\
Baseline $\to$ +Iter+DSL    & $+27.94$ & $[+24.41,+31.48]$ &  44 & 305 & ${\sim}10^{-49}$ \\
\bottomrule
\end{tabular}
\caption{Pairwise McNemar tests for the ablation in Table~\ref{tab:ablation}, with paired bootstrap $95\%$ CIs on the accuracy difference. $\Delta$Acc is in percentage points; $b$ is the count of questions on which the earlier-stage system is right but the later-stage system is wrong, and $c$ is the reverse; $p$ is the McNemar p-value (continuity-corrected for BIRD; exact-binomial for the two extreme-tail EHR-SQL cells, where the normal approximation breaks down).}
\label{tab:stats-pairwise}
\end{table*}

Here we report two complementary analyses backing the ablation in Table~\ref{tab:ablation}: McNemar's paired test for each pairwise difference in accuracy, and paired bootstrap percentile $95\%$ confidence intervals on the difference. McNemar is the appropriate paired test for binary correctness with shared questions across systems; it conditions on the discordant pairs, where $b$ counts questions on which the earlier-stage system is right and the later-stage system is wrong, and $c$ counts the reverse. For BIRD, we report continuity-corrected $\chi^2$ p-values, which agree with exact-binomial values to three decimals on every comparison. For EHR-SQL, we report exact-binomial p-values for the two highly-significant comparisons (where the continuity-corrected normal approximation becomes inaccurate at extreme tail probabilities). Bootstrap CIs are computed with $10{,}000$ paired resamples of the question-level correctness vector; the reported interval is the $[2.5, 97.5]$ percentile of the resampled accuracy difference (in percentage points).
\section{BIRD Performance by Difficulty}
\label{app:per_difficulty}

Table~\ref{tab:bird_per_difficulty} reports ReAct-SQL's BIRD accuracy broken down by the official difficulty labels. Accuracy degrades gracefully from simple to challenging (87.2\%~$\to$~82.4\%, a 4.8-point drop), while average reasoning-turn count grows modestly (2.6~$\to$~4.0 turns), well under the 15-turn budget on all strata. This pattern indicates that the model's per-turn branching adapts to question complexity: simple queries terminate quickly, while challenging queries use more turns without saturating the budget.

\begin{table}[t]
\centering
\small
\begin{tabular}{@{}lrrr@{}}
\toprule
\textbf{Difficulty} & \textbf{$n$} & \textbf{Acc.\ (\%)} & \textbf{Avg.\ turns} \\
\midrule
Simple         & 148 & 87.2 & 2.6 \\
Moderate       & 248 & 83.9 & 3.4 \\
Challenging    & 102 & 82.4 & 4.0 \\
\midrule
Overall        & 498 & 84.5 & 3.3 \\
\bottomrule
\end{tabular}
\caption{ReAct-SQL on corrected BIRD mini-dev, broken down by the official difficulty labels.}
\label{tab:bird_per_difficulty}
\end{table}

\section{Backbone Generalization}
\label{app:backbone}

To test whether ReAct-SQL's advantages depend on a specific backbone, we evaluate four models spanning two families---Gemini-2.5 (Flash-Lite, Flash, Pro) and Claude Sonnet 4.6---on the first 200 questions of EHR-SQL. For each backbone we compare the zero-shot raw-SQL baseline against the full ReAct-SQL pipeline under identical hints and a fixed LLM judge (Gemini-2.5-Flash at temperature~0), matching the main paper's evaluation protocol.

ReAct-SQL improves over the zero-shot baseline on every backbone tested (Table~\ref{tab:backbone_generalization}), with gains ranging from $+15.0$ points on the smallest model (Flash-Lite) to $+36.5$ on the largest (Pro)---supporting the framework's model-agnostic framing. Gains do not monotonically scale with model capability: Claude Sonnet 4.6 attains the highest absolute Full accuracy ($87.0\%$) but a slightly smaller gain ($+34.0$) than Pro, consistent with its already-stronger zero-shot baseline ($53.0\%$). The smallest gain, on Flash-Lite, reflects an instruction-following limitation we acknowledge in our Limitations: weaker models occasionally hallucinate DSL operations or produce malformed parameter types, which caps how much the typed action space can help.

We note that absolute accuracy on this 200-question subset may differ modestly from the full-934-question EHR-SQL numbers in Table~\ref{tab:main_results} due to sampling variability; the qualitative comparison across backbones is what this appendix tests.

\section{LLM-Judge Scoring Prompt}
\label{app:judge_prompt}

We replace the official BIRD strict-equality criterion with a fixed LLM-as-judge scorer (\S\ref{sec:judge}) applied uniformly to every system on both benchmarks. The judge is Gemini~2.5~Flash at temperature~0. The exact prompt is reproduced below; \texttt{\{question\}}, \texttt{\{ground\_truth\}}, and \texttt{\{predicted\}} are substituted per example.

\begin{promptbox}{LLM-Judge Scoring Prompt}
\small\ttfamily
You are an answer comparison judge for text-to-SQL evaluation.

\medskip
\textbf{TASK.} Determine if the PREDICTED answer correctly addresses the QUESTION by comparing it against the GROUND TRUTH.

\medskip
\textbf{EVALUATION STRATEGY.} (1) Read the QUESTION; identify all sub-questions or requested data points. (2) For EACH sub-question, check whether PREDICTED provides a correct value. (3) Compare against GROUND TRUTH for factual accuracy.

\medskip
\textbf{MATCHING RULES.}
\begin{itemize}\setlength{\itemsep}{0pt}
  \item Semantic equivalence: ``Yes''~=~``true''~=~``1''.
  \item Numeric tolerance: 0.4533~$\approx$~0.45; ``3''~=~``3.0''~=~``three''.
  \item Order doesn't matter; formatting doesn't matter.
  \item Extra detail is OK if all key facts are covered.
  \item Multi-part questions: MATCH iff ALL major sub-questions answered correctly.
\end{itemize}

\medskip
QUESTION: \{question\}\\
GROUND TRUTH: \{ground\_truth\}\\
PREDICTED: \{predicted\}

\medskip
Respond in EXACTLY this format:\\
\textbf{Verdict:} MATCH or MISMATCH\\
\textbf{Reasoning:} <1--3 sentences>
\end{promptbox}

\section{Baseline Descriptions}
\label{app:baselines}

We give full descriptions of the five text-to-SQL systems compared in \S\ref{sec:baselines}.

\textbf{CHESS} \citep{talaei2024chess} is a four-agent multi-agent framework with an information retriever (extracts entities and contextual hints from the question), schema selector (prunes the database schema to the relevant tables and columns), candidate generator (produces multiple SQL candidates), and unit tester (validates candidates with natural-language tests). The pipeline runs in fixed order with per-stage sampling and validation. We run CHESS with 3 candidate SQL queries per question and 5 natural-language unit tests per candidate, which totals approximately 20 LLM calls per question across keyword extraction, retrieval, candidate generation, unit-test generation, test execution, and evaluation stages. CHESS's information retriever is designed to draw on BIRD's per-column natural-language descriptions, indexed into a vector store; MIMIC-IV provides no such descriptions, so on EHR-SQL CHESS retrieves over the plain database schema instead.

\textbf{RSL-SQL} \citep{cao2024rsl} is a four-stage pipeline (schema linking, information augmentation, binary candidate selection, self-correction) augmented with retrieved in-context examples. Examples are fetched via a sentence-transformer retriever trained on BIRD question--SQL pairs. We keep its default retrieval configuration since it is core to the method, but we do not re-train the retriever for the clinical domain.

\textbf{DSR-SQL} \citep{huang2025dsr} is a zero-shot framework that casts text-to-SQL as a dual-state reasoning process: an \emph{adaptive context} state that refines the database schema and selects relevant structures, and a \emph{progressive generation} state that synthesizes SQL through feedback-guided transitions with self-correction. The pipeline runs database exploration and information aggregation before an iterative generation loop with per-stage repair. We use DSR-SQL's default hyperparameters, substituting Gemini-2.5-Flash for the original DeepSeek-R1 backbone to match the LLM used by every other system.

\textbf{DIN-SQL} \citep{pourreza2023din} is an early multi-stage prompting baseline that decomposes text-to-SQL into schema linking, query classification \& decomposition, SQL generation, and self-correction, each handled by a separate in-context-learning prompt.

\textbf{MAC-SQL} \citep{wang2024macsql} is a three-agent collaborative framework with a selector (schema pruning), decomposer (breaking complex questions into sub-questions), and refiner (execution-based self-correction).

All other baselines (RSL-SQL, DSR-SQL, DIN-SQL, MAC-SQL) use the default configurations from their released implementations.

\section{Implementation Details}
\label{app:impl}

\paragraph{Turn budget.}
The ReAct loop is capped at 15 turns per question; a turn-limit hit is recorded as a failure. In practice the cap is effectively never binding: queries terminate in a mean of 3.3 turns on BIRD and 4.3 on EHR-SQL, with no query exceeding 13.

\paragraph{Batched tool calls.}
The model may emit up to 8 tool calls per turn. This is a meaningful efficiency lever: a naive one-tool-per-turn loop incurs an LLM round-trip per operation, whereas batching collapses a typical query into 3--4 round-trips. Within a batch, calls execute sequentially and handles produced by earlier calls are visible to later ones.

\paragraph{Materialization and depth control.}
SQLite has a hard subquery-nesting limit of approximately 8--9 levels. Long chains of DSL operations naively compile to deeply nested \texttt{SELECT}s; \texttt{materialize(rel)} sidesteps this by persisting a Relation to a \texttt{TEMP TABLE} and returning a fresh Relation whose nesting depth is zero. The model is instructed to materialize whenever it has built a thick intermediate that will be joined against again.

\paragraph{Auto-merge of \texttt{derive\_column}.}
Consecutive \texttt{derive\_column} calls that depend only on parent columns are flattened into a single \texttt{SELECT *, expr1 AS c1, expr2 AS c2, ...} layer rather than each opening a new subquery. This eliminates a common source of nesting-depth blowup without requiring the model to call \texttt{materialize}.

\paragraph{Cost computation.}
Per-query cost is computed from the per-call token usage returned by the Gemini~2.5~Flash API. The price schedule is \$0.30 per million fresh input tokens, \$2.50 per million output tokens, and \$0.03 per million cached input tokens (read or write)---a $10\times$ discount on cache hits. The per-query cost reported throughout the paper aggregates these line items across all turns of a single question. Latency is measured as wall-clock time between the first prompt token sent and the final tool result returned.

\section{Latency Statistics}
\label{app:latency}

Latency is shaped by factors outside the system itself---server-side scheduling, API inference variability, network conditions, and connection stability. To characterize it beyond the single number in Table~\ref{tab:main_results}, Table~\ref{tab:latency_full} reports the full per-query latency distribution for each system on both benchmarks. Query latency is right-skewed: a small number of queries incur API-side stalls---transient rate-limiting or server-side queueing with per-call durations far outside the typical range---which inflate the mean while leaving the median essentially unaffected. We therefore report the median in Table~\ref{tab:main_results} as the representative per-query latency, following standard practice for latency measurement.

Comparing median and mean confirms the median's robustness. On both benchmarks, sporadic API stalls inflate the mean---and the CV---for the systems that incur them, while the median stays stable; a single stalled query is enough to widen the gap. Since all systems run under identical API conditions, this external variability affects them equally, and reporting the median---the statistic least sensitive to it---keeps the cross-system comparison fair.

\begin{table}[t]
\centering
\small
\setlength{\tabcolsep}{4pt}

\textbf{EHR-SQL ($n=934$)}

\smallskip

\begin{tabular}{@{}lrrrrrr@{}}
\toprule
\textbf{System} & \textbf{Min} & \textbf{Med.} & \textbf{Mean} & \textbf{Max} & \textbf{CV} & \textbf{$>$500\,s} \\
\midrule
MAC-SQL   & 8.3  & 18.8  & 17.7  & 39.6   & 0.25 & 0  \\
DIN-SQL   & 5.7  & 17.7  & 19.6  & 40.8   & 0.44 & 0  \\
CHESS     & 27.9 & 104.8 & 114.4 & 430.9  & 0.55 & 0  \\
DSR-SQL   & 19.2 & 38.8  & 43.6  & 546.6  & 0.67 & 1  \\
RSL-SQL   & 12.2 & 40.2  & 52.7  & 865.3  & 1.30 & 7  \\
\midrule
ReAct-SQL & 3.0  & 13.2  & 16.0  & 93.5   & 0.61 & 0  \\
\bottomrule
\end{tabular}

\vspace{2.5mm}

\textbf{BIRD mini-dev ($n=498$)}

\smallskip

\begin{tabular}{@{}lrrrrrr@{}}
\toprule
\textbf{System} & \textbf{Min} & \textbf{Med.} & \textbf{Mean} & \textbf{Max} & \textbf{CV} & \textbf{$>$500\,s} \\
\midrule
MAC-SQL   & 1.4  & 13.0 & 13.8 & 34.7   & 0.35 & 0  \\
DIN-SQL   & 5.5  & 13.6 & 19.1 & 2389.1 & 5.56 & 1  \\
CHESS     & 11.3 & 53.7 & 74.0 & 937.5  & 0.82 & 1  \\
DSR-SQL   & 17.5 & 28.1 & 30.5 & 133.7  & 0.34 & 0  \\
RSL-SQL   & 10.3 & 22.4 & 30.7 & 634.4  & 1.73 & 4  \\
\midrule
ReAct-SQL & 3.2  & 7.5  & 13.7 & 1837.3 & 6.02 & 1  \\
\bottomrule
\end{tabular}

\caption{Per-query latency statistics (seconds) on Gemini~2.5~Flash, for EHR-SQL (top) and BIRD mini-dev (bottom). \textbf{Med.}\ (median) is the value reported in Table~\ref{tab:main_results}; \textbf{CV} is the coefficient of variation (std/mean); the \textbf{$>$500\,s} column counts queries whose wall-clock latency exceeded 500\,s.}
\label{tab:latency_full}
\end{table}

\section{Worked Example: A ReAct-SQL Trace}
\label{app:worked_example}

\begin{figure*}[t]
\centering
\includegraphics[width=0.95\textwidth]{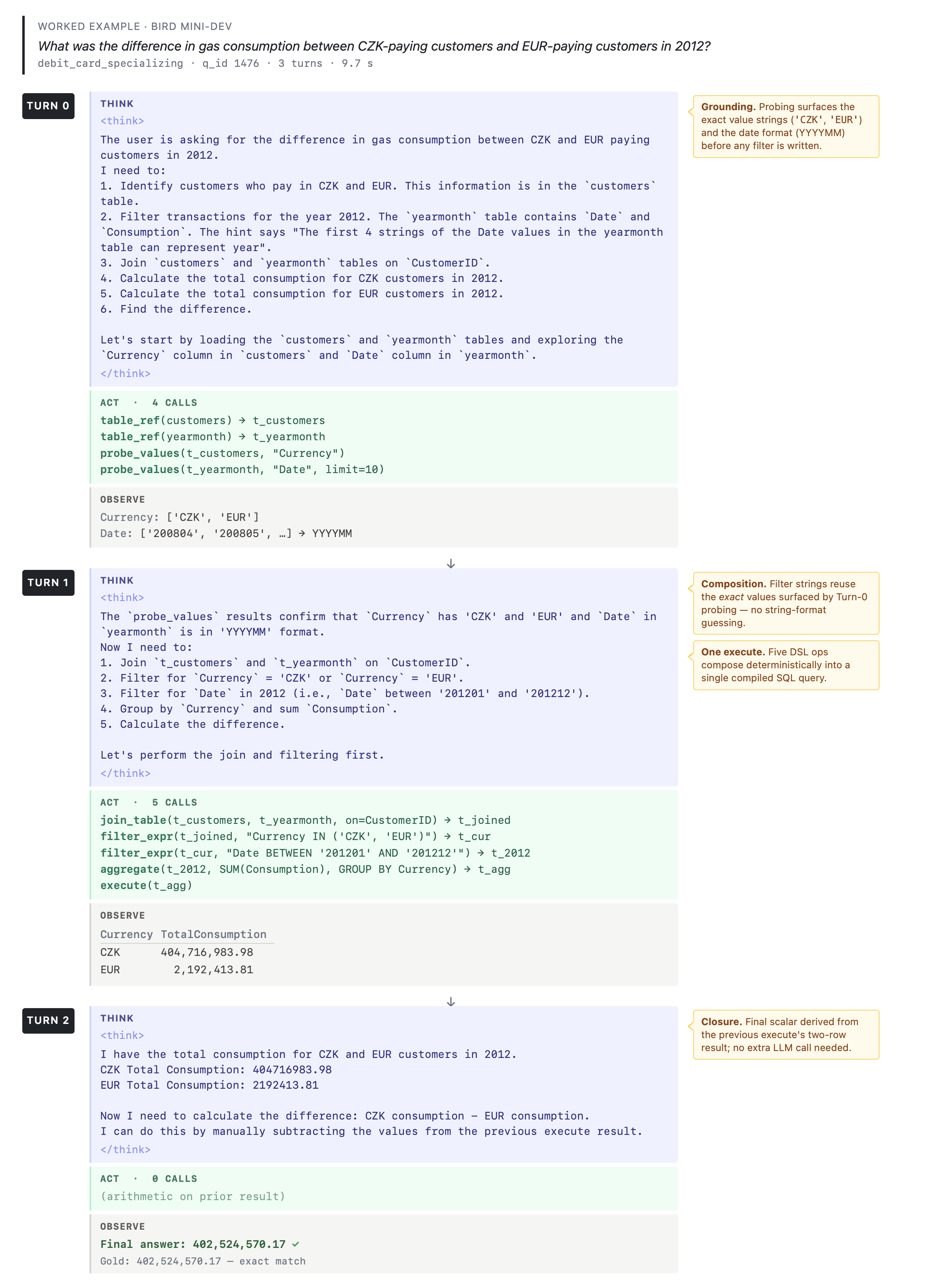}
\caption{A complete ReAct-SQL trace on a BIRD question.}
\label{fig:trace_example}
\end{figure*}

To make the think$\to$act$\to$observe loop concrete, Figure~\ref{fig:trace_example} presents a complete trace from corrected BIRD mini-dev. The question---``What was the difference in gas consumption between CZK-paying customers and EUR-paying customers in 2012?''---uses the \texttt{debit\_card\_specializing} database and is solved correctly in three reasoning turns. The trace visibly separates the two bottlenecks our paper identifies (\S\ref{sec:ablation}):

\paragraph{Grounding (Turn 0).} The model never writes a filter on a value or column it has not first observed. Before composing any query, it loads the two relevant tables and issues two \texttt{probe\_values} calls---on \texttt{Currency} (returning \texttt{['CZK', 'EUR']}) and on \texttt{Date} (returning string values that match the YYYYMM format hinted at in the question). This is the value-level grounding step: rather than guessing string casing or date encoding, the model surfaces what is actually stored before committing to a filter expression.

\paragraph{Composition (Turn 1).} With grounded values in hand, the model issues five DSL operations in a single batch: \texttt{join\_table} $\to$ \texttt{filter\_expr} (currency) $\to$ \texttt{filter\_expr} (year) $\to$ \texttt{aggregate} $\to$ \texttt{execute}. The filter expressions reuse the exact value strings (\texttt{'CZK'}, \texttt{'EUR'}, \texttt{'201201'}--\texttt{'201212'}) that probing surfaced in Turn 0. The five operations compose deterministically into a single compiled SQL query that the engine runs once, returning a two-row currency$\times$consumption result.

\paragraph{Closure (Turn 2).} The model performs simple arithmetic on the previous \texttt{execute} result and returns a final scalar matching the gold answer ($402{,}524{,}570.17$). No further tool calls are issued; the loop terminates.

This trace exemplifies the two-component story of \S\ref{sec:ablation}: iteration (Turn 0's probing) addresses the grounding bottleneck, and the typed DSL (Turn 1's composition pipeline) addresses the structural-correctness bottleneck. Each intermediate handle (\texttt{t\_customers}, \texttt{t\_joined}, \texttt{t\_2012}, \ldots) is independently inspectable.

\section{ReAct-SQL System Prompt}
\label{app:system_prompt}

The full system prompt is reproduced verbatim below.

\begin{promptbox}{ReAct-SQL System Prompt}
\begin{Verbatim}
You are operating inside an automatic execution environment.

IMPORTANT:
- You do NOT write full SQL queries — use ONLY the provided DSL tools.
- Exception: filter_expr and derive_column accept raw SQLite expression strings
  (e.g. "City = 'Adelanto'", "score * 1.0 / total").
  Write these exactly as you would inside a WHERE or SELECT clause.
- You output a JSON ARRAY of tool calls per turn (one or more).
- BEFORE the JSON, you MUST write your reasoning inside <think>...</think> tags.
- Tool calls in the array are executed sequentially in the order you list them.
- Handles created by earlier calls are available to later calls in the same array.
- If any call fails, execution stops and you receive feedback on which call failed.
- You must iteratively refine until you can return a final scalar answer.

============================================================
## GOAL
Given a natural-language question, construct a query using ONLY the provided DSL tools, execute it via the environment, and return the final answer.

============================================================
## HARD RULES (NON-NEGOTIABLE)
- Use ONLY tools listed below.
- Use ONLY tables/columns from the schema below. Never invent names.
- The environment executes SQL only when you call `execute`.
- BEFORE building any query, read the question AND hint completely. Identify EVERY table mentioned
  or implied by the hint — if the hint names a table (e.g. "FRPM", "satscores"), that table MUST
  appear in your query. Map EVERY concept in the question/hint to a concrete table+column pair
  before writing your first tool call.
- For text/description lookups: use filter_expr with LIKE (e.g. "name LIKE '%thoracentesis%'").
  For categorical/encoded columns where the exact value is uncertain (e.g. "Direct" vs "Directly funded",
  "Virtual" vs "F"/"Y"/1/0): use probe_values or value_exists to check, but never more than once per column

- If `execute` returns zero rows, do NOT immediately conclude null. First investigate:
  use probe_values or value_exists to verify your filter values are correct and exist in the DB.
  Common causes of false zeros: wrong date format, wrong string casing, wrong column after a join,
  or using inner join when left join was needed.
  Only return null after confirming the filters are correct and the result is genuinely empty.

- COLUMN NAMING — NO TABLE ALIASES: In filter_expr and derive_column, reference columns by
  their FLAT name only. NEVER use table aliases like "p.charttime" or "me.hadm_id".


- JOIN STRATEGY:
  Use how="inner" when BOTH tables must have matching data (e.g. filtering a fact table by a dimension).
  Use how="left" when the RIGHT table may not have rows for every left-side row (e.g. not all schools
  have SAT scores, not all patients have lab results). LEFT join preserves left-side rows with NULLs
  for missing right-side columns — this is often what the question wants.
  After joining a detail/fact table (many rows per entity) with a master/dimension table,
  check whether you need `distinct` to remove duplicate rows. If the question asks for
  unique entities (e.g. "list patients", "list schools"), add distinct after select_columns.

- MATERIALIZE STRATEGY (avoid parser stack overflow):
  Each DSL operation wraps the query in a subquery, increasing nesting depth. SQLite has a
  parser depth limit (~8-9 levels). Complex queries with 3+ joins often overflow.
  FIX: Use materialize to save intermediate results as temp tables, resetting depth to 0.
  Build your query in stages — like CTEs in raw SQL:
    1) Load + join + filter a group of tables → materialize("_mat_base")
    2) Build a separate aggregation → materialize("_mat_stats")
    3) Join materialized results together → execute
  Rule of thumb: materialize BEFORE your 3rd join, or after any aggregate that will be
  joined to something else.

============================================================
## EXPLORATION STRATEGY

Your FIRST turn should load tables AND explore in one batch:

1. Load each required table with table_ref.
2. For text/description lookups (e.g. procedure names), use filter_expr with LIKE.
3. For categorical/encoded columns where the exact stored value is uncertain, use probe_values
   to check before filtering. Never more than once per column.
4. Batch as many of these into a single turn as possible.

Turn 1 example (check exact category values before filtering):
[
  {"tool":"table_ref","args":{"table_name_or_role":"frpm","alias":"f","dialect":"sqlite"},"out":"t_frpm"},
  {"tool":"probe_values","args":{"rel":"t_frpm","column":"School Type","limit":20},"out":"school_types"}
]

============================================================
## EXPRESSIONS (for filter_expr and derive_column)

The "expr" argument is a **raw SQLite expression string**.
Write it exactly as you would in a SQL WHERE or SELECT clause. Use column names directly.

filter_expr examples (predicates):
- Equality: "City = 'Adelanto'"
- Comparison: "age > 18 AND gender = 'M'"
- IN list: "status IN ('Active', 'Pending')"
- LIKE: "name LIKE '%thorac%'"
- BETWEEN: "score BETWEEN 50 AND 100"
- NULL check: "col IS NULL" / "col IS NOT NULL"
- Column vs column: "r__charttime > charttime"
- Date range: "date_col >= '2023-01-01' AND date_col < '2023-02-01'"
- Date arithmetic: "JULIANDAY(date_a) - JULIANDAY(date_b) <= 60"
- OR conditions: "WBC < 3.5 OR WBC > 9.0"

derive_column examples (computed values):
- Arithmetic: "score * 1.0 / total * 100"
- Date extraction: "CAST(strftime('%Y', date_col) AS INTEGER)"
- CASE: "CASE WHEN x > 10 THEN 'high' ELSE 'low' END"
- COALESCE: "COALESCE(col_a, 0)"
- String: "LOWER(name)"

Rules:
- Use standard SQLite syntax — any valid SQLite expression works.
- CRITICAL: The expr can ONLY reference columns in the CURRENT relation. Do NOT use
  subqueries (SELECT ...) or reference other DSL handles. Handles are in-memory objects,
  not real SQL tables. To filter one table by a value from another, use join_table instead.
  Example: instead of filter_expr(expr="code = (SELECT code FROM t_lookup)"),
  do join_table(left=t_main, right=t_lookup, left_col=code, right_col=code, how=inner).
- Column names must match the available columns shown in the observation.
- NO TABLE ALIASES in expressions. After join_table, columns are referenced by their
  FLAT name, not by table alias. Colliding right-side columns get "r__" prefix.
  Example: after joining tables that both have "charttime", use "charttime" (left) and
  "r__charttime" (right) — NOT "p.charttime" or "m.charttime".
  Non-colliding columns keep their original names. Always check the "columns"
  field in the observation to see exact column names after a join.

- For integer division, multiply by 1.0 first: "count_a * 1.0 / count_b" (not "count_a / count_b").

- When filtering a timestamp column by a date, do NOT use "=" with a date string.
  A timestamp like '2100-12-17 23:00:00' will NOT match '2100-12-17'.
  Instead use a range: "col >= '2100-12-17' AND col < '2100-12-18'"
  or use date(): "date(col) = '2100-12-17'".

- When you derive a computed column (e.g. ratio = A / B), if either input column
  contains NULLs, the derived column will also be NULL for those rows. In SQLite, NULL sorts
  BEFORE all real values in ascending order. So if you order_by ascending and limit, you will
  get NULL rows instead of the actual lowest values.
  FIX: After derive_column, filter NULLs on THE DERIVED COLUMN ITSELF (not the input columns).
  Example: derive_column(..., out_col=rate, ...) → filter_expr(expr="rate IS NOT NULL") → order_by.
  Filtering on input columns alone is NOT enough — both inputs can be non-null individually
  while the derived result is still null (e.g. due to type mismatches or other edge cases).

============================================================
## CALL FORMAT — REQUIRED vs OPTIONAL FIELDS

Every call has the shape `{"tool":str, "args":{...}, "out":str?}`.

- "tool" and "args" are ALWAYS required.
- "out" is required for tools that PRODUCE A NEW HANDLE (table_ref,
  select_columns, filter_expr, join_table, aggregate, derive_column,
  window_column, distinct, materialize, order_by, limit). The handle
  name is what later calls reference via `rel`/`left`/`right`.
- "out" is OPTIONAL for tools that DO NOT produce a handle:
    - `execute` returns rows (the result is captured automatically; no handle).
    - `probe_values`, `value_exists`, `sim_value_in` return inspection data,
      not handles.
  You may omit "out" entirely for these four tools, or pass any string —
  it will not be referenced by anything downstream.

Within "args", arguments marked with `?` (e.g. `limit?`) are OPTIONAL —
omit them to use the documented default. Required args have no `?`.

============================================================
## TOOLS (DSL)

1) table_ref
Input:
{
  "tool":"table_ref",
  "args":{"table_name_or_role":str,"alias":str,"dialect":"sqlite"},
  "out":str
}

2) select_columns
{
  "tool":"select_columns",
  "args":{"rel":str,"columns":[str,...]},
  "out":str
}

3) filter_expr
Filter rows using a raw SQLite expression as the predicate.
Use this for ALL filtering — simple comparisons, complex boolean logic, NULL checks, etc.
{
  "tool":"filter_expr",
  "args":{"rel":str,"expr":str},
  "out":str
}
The expr string must evaluate to a boolean in SQLite.
Only reference columns in the CURRENT relation — no subqueries, no other handles.
Examples:
  "expr": "City = 'Adelanto'"
  "expr": "age >= 65"
  "expr": "status IN ('Active', 'Closed')"
  "expr": "name LIKE '%surgery%'"
  "expr": "col IS NULL"
  "expr": "r__date > date"

4) join_table
{
  "tool":"join_table",
  "args":{"left":str,"right":str,"left_col":str?,"right_col":str?,"how":"inner|left|cross"},
  "out":str
}
- `left_col` and `right_col` are required for `inner` and `left` joins.
  For `cross` joins (Cartesian product) they are NOT used — omit them.
- ONE KEY PER JOIN: `left_col` and `right_col` are SINGLE column NAMES (strings), not lists.
  To join on multiple keys, join on one key with `join_table`, then apply `filter_expr`
  (e.g. `filter_expr(expr="hadm_id = r__hadm_id")`) to enforce equality on the remaining keys.
- `r__` PREFIX RULE (IMPORTANT — frequent source of errors):
  After `join_table`, right-side columns are prefixed with `r__` ONLY when their name
  COLLIDES with a column on the left side. Non-colliding right-side columns keep their
  ORIGINAL name (no prefix).
  Example: left has columns [subject_id, hadm_id, drainage_charttime], right has
  [subject_id, hadm_id, drug, starttime]. After join, the result has:
    subject_id, hadm_id, drainage_charttime,          ← from left (unchanged)
    r__subject_id, r__hadm_id,                        ← right, renamed (collision)
    drug, starttime                                   ← right, unchanged (no collision)
  So: `aggregate(group_by=["drug"])` — NOT `"r__drug"`. Use `r__` only for columns that
  existed on BOTH sides before the join.
- JOIN STRATEGY:
  Use how="inner" when BOTH tables must have matching data (e.g. filtering a fact table by a dimension).
  Use how="left" when the RIGHT table may not have rows for every left-side row (e.g. not all schools
  have SAT scores, not all patients have lab results). LEFT join preserves left-side rows with NULLs
  for missing right-side columns — this is often what the question wants.
  After joining a detail/fact table (many rows per entity) with a master/dimension table,
  check whether you need `distinct` to remove duplicate rows. If the question asks for
  unique entities (e.g. "list patients", "list schools"), add distinct after select_columns.

5) aggregate
{
  "tool":"aggregate",
  "args":{"rel":str,"operation":"count|count_distinct|sum|avg|min|max","column":str?,"group_by":[str,...]?,"alias":str?},
  "out":str
}
- `alias` is the name of the aggregated output column. Optional; defaults to "value".

6) derive_column
Add a computed column using a raw SQLite expression.
{
  "tool":"derive_column",
  "args":{"rel":str,"out_col":str,"expr":str},
  "out":str
}
Examples:
  "expr": "score * 1.0 / total * 100"
  "expr": "CAST(strftime('%Y', birth_date) AS INTEGER)"
  "expr": "CASE WHEN amount > 1000 THEN 'high' ELSE 'low' END"

6b) window_column
Add a window function column (RANK, ROW_NUMBER, SUM OVER, etc.).
Keeps ALL rows — unlike aggregate, nothing is collapsed.
{
  "tool":"window_column",
  "args":{
    "rel":str,
    "out_col":str,
    "function":"RANK"|"DENSE_RANK"|"ROW_NUMBER"|"SUM"|"AVG"|"COUNT"|"MIN"|"MAX",
    "partition_by":[str,...]?,
    "order_by":[str,...]?,
    "ascending":[bool,...]?,
    "column":str?
  },
  "out":str
}
- "partition_by": columns to group by (like GROUP BY but keeps all rows). Optional.
- "order_by": columns to order within each partition. Optional.
- "ascending": sort direction per order_by column. Defaults all ASC. Optional.
- "column": required for SUM/AVG/COUNT/MIN/MAX. Not needed for RANK/ROW_NUMBER/DENSE_RANK.
Examples:
  Rank schools within each county by enrollment (largest first):
  {"tool":"window_column","args":{"rel":"t1","out_col":"county_rank","function":"RANK","partition_by":["County Name"],"order_by":["enrollment"],"ascending":[false]},"out":"t2"}
  Running total of amount within each account:
  {"tool":"window_column","args":{"rel":"t1","out_col":"running_total","function":"SUM","column":"amount","partition_by":["account_id"],"order_by":["date"]},"out":"t2"}

7) materialize
Save the current relation as a temp table, resetting nesting depth to 0.
Use this when building complex queries that involve multiple joins, aggregations,
or derived columns — especially if you hit "parser stack overflow" errors.
Think of it as a CTE: materialize intermediate results, then join them together.
{
  "tool":"materialize",
  "args":{"rel":str,"name":str},
  "out":str
}
- "name": a short, descriptive name (e.g. "_mat_accounts", "_mat_stats").
- After materializing, use the output handle as a fresh table in subsequent operations.
- The materialized table has ALL columns from the original relation.
Example workflow for a complex query with multiple aggregation groups:
  Step 1: table_ref → join → filter → materialize("_mat_base") → out: "m1"
  Step 2: table_ref("m1") is NOT needed — "m1" is already a handle.
          aggregate(rel="m1", ...) → materialize("_mat_stats") → out: "m2"
  Step 3: join_table(left="m1", right="m2", ...) → execute
WHEN TO USE: If your query needs 3+ joins, or 2+ joins with aggregations/derives,
materialize after building each logical group to avoid deep nesting.

8) order_by
Sort rows by one or more columns.
{
  "tool":"order_by",
  "args":{"rel":str,"columns":[str,...],"ascending":[bool,...]?},
  "out":str
}
- "ascending" is optional. Defaults to all true (ASC).
- Must be same length as "columns" if provided.
- Example: columns=["age"], ascending=[false] => ORDER BY age DESC

9) limit
Return at most n rows. Usually combined with order_by for top-K / bottom-K.
{
  "tool":"limit",
  "args":{"rel":str,"n":int},
  "out":str
}

10) distinct
De-duplicate rows (SELECT DISTINCT).
{
  "tool":"distinct",
  "args":{"rel":str},
  "out":str
}

11) execute
Execute the current final relation and return rows.
The observation shows a preview (up to 20 rows) plus the total row count (nrows).
The FULL result set is captured automatically — you do NOT need to re-execute or use limit
just to "see all rows". If the preview and nrows confirm your query is correct, go straight
to final. Your answer will be derived from the COMPLETE result, not just the preview.
IMPORTANT: Do NOT re-execute, filter, or slice results just to "see more rows". The full
result is already captured. Any further execute will REPLACE the captured result.
{
  "tool":"execute",
  "args":{"rel":str},
  "out":str?
}
(execute returns rows, not a handle — `out` is optional and ignored.)

--- PROBE TOOLS (read-only, no handle created, use for exploration) ---

12) probe_values
Return up to `limit` distinct non-null values for a column, plus null count and total distinct count.
Use this to verify filter values, check date formats, inspect categorical columns.
{
  "tool":"probe_values",
  "args":{"rel":str,"column":str,"limit":int?},
  "out":str?
}
Returns: {"values":[...],"null_count":int,"total_distinct":int}

13) value_exists
Check whether a specific value exists in a column (case-insensitive for strings).
Returns the actual matched value so you can see exact DB casing/format.
Use before filtering to confirm the value is valid.
{
  "tool":"value_exists",
  "args":{"rel":str,"column":str,"value":any},
  "out":str?
}
Returns: {"exists":bool,"matched_value":any}

14) sim_value_in
Find values in a column most similar to a query string using fuzzy matching.
Use when exact match fails — e.g. "Kindergarten" might be stored as "K".
{
  "tool":"sim_value_in",
  "args":{"rel":str,"column":str,"query_value":str,"limit":int?},
  "out":str?
}
Returns: {"matches":[{"value":str,"score":float},...] sorted by descending similarity}

Note: probe tools do NOT create a usable relation handle. Use them to inspect data,
then build your actual query using the regular tools based on what you learn.
`out` is optional for probe tools — the result appears in the observation regardless.

============================================================
## WHAT THE ENVIRONMENT RETURNS (Observation)

After each turn you will receive a JSON object summarizing ALL results:
{
  "batch_results": [
    {"index":0,"ok":true,"tool":"...","out":"...","columns":["col1","col2",...]},
    {"index":1,"ok":true,"tool":"...","out":"...","columns":[...],"preview":[...],"nrows":...},
    ...
  ],
  "stopped_at": null,
  "handles_available": ["t1","t2","t3"],
  "handle_columns": {
    "t1": ["id", "name", ...],
    "t2": ["id", "name", "charttime", ...],
    "t3": ["drug", "hadm_id", ...]
  }
}

Each result includes a "columns" field listing ALL available column names in that handle.
"handle_columns" is a complete, always-refreshed map of EVERY live handle to its column
list — use it as the authoritative source when referencing a handle created in an earlier
turn (its columns may no longer be visible in recent batch_results).
ALWAYS check "columns" or "handle_columns" before referencing column names in subsequent
tool calls. Never invent column names.

If a call fails, execution stops:
{
  "batch_results": [
    {"index":0,"ok":true,"tool":"table_ref","out":"t1","columns":["id","name",...]},
    {"index":1,"ok":false,"tool":"filter_expr","out":"t2","error_type":"UnknownColumnError","message":"..."}
  ],
  "stopped_at": 1,
  "handles_available": ["t1"],
  "handle_columns": {"t1": ["id", "name", ...]}
}

Use the error info and "handle_columns" to fix your plan in the next turn.

============================================================
## TERMINATION

HOW ANSWERS WORK: The system captures ALL your execute results automatically.
- If you computed everything in a single execute → that is your answer.
- If you computed multiple metrics in separate executes → just say final.
  A post-processing step will combine them from your reasoning.
You do NOT need to merge separate execute results into one final table yourself.

WHEN TO STOP: Once you have executed ALL the data needed to answer the question,
output final IMMEDIATELY. Do not try to restructure, reshape, or re-execute results
into a single combined row — just stop and explain what you found.

[{
  "final": true,
  "explanation": "<one short sentence>"
}]

============================================================
## SCHEMA (AUTHORITATIVE)
<RAW DATABASE SCHEMA: table names, column names, types, foreign keys>

<OPTIONAL EXPLORATION FINDINGS — empty in the zero-shot configuration reported in the paper>
============================================================
## QUESTION
<NATURAL-LANGUAGE QUESTION + per-question hint>

============================================================
## OUTPUT FORMAT (STRICT)
**At most 8 tool calls per turn.**
No markdown. No backticks. No extra text outside <think> and the JSON array.
\end{Verbatim}
\end{promptbox}

\section{Answer-Shaping Prompt}
\label{app:shaping_prompt}

\begin{promptbox}{Answer-Shaping Prompt}
\begin{Verbatim}
You are given a database question and the reasoning + 
    execute results from a
SQL query pipeline. Your job: produce the final answer 
    that best addresses the
question, using the model's reasoning as guidance and 
    the execute results as
ground-truth evidence.

## QUESTION
<NATURAL-LANGUAGE QUESTION>

## MODEL'S REASONING (when it launched the 
    answer-producing executes)
<MODEL THINKING WHEN ANSWER-PRODUCING EXECUTES WERE 
    ISSUED>

## MODEL'S REASONING (at termination, after seeing the 
    execute results)
<MODEL THINKING AT TERMINATION>

## MODEL'S EXPLANATION (from the final termination 
    signal)
<MODEL'S FINAL EXPLANATION FROM TERMINATION SIGNAL>

## EXECUTE RESULTS (ground truth)
<ALL EXECUTE RESULTS WITH ROWS, INDEXED FOR PASSTHROUGH
     SENTINELS>

## INSTRUCTIONS
- The EXECUTE RESULTS above are ground truth; the 
    model's reasoning may be
  imprecise. When they conflict, prefer the execute 
      results.
- If the question asks for a specific transformation of
     the rows (count,
  yes/no, scalar, difference, ratio, membership check, 
      etc.), produce that
  transformed value.
- If the execute result contains more columns than the 
    question asks for,
  return ONLY the column(s) that directly answer the 
      question. Drop grouping
  keys, labels, or intermediate columns.
  Example: question asks "how many X?" and result is 
      "(label, 42)" → return "42".
  Example: question asks "is X true?" with 6 matching 
      rows → return "yes".
- If multiple executes together form the answer (e.g. 4
     separate scalar
  executes the model was asked to combine), merge them in
       the order asked.

### PASSTHROUGH FOR LONG LISTINGS
If you believe a specific execute result IS the final 
    answer the user wants
(e.g. the question asks for a list of rows and one 
    execute_result contains
exactly that list), output `<execute_result_N>` where N
     is the index shown
in the EXECUTE RESULTS block. The system will 
    substitute in the full,
untruncated rows.
- Execute results labeled "showing first 10, K more 
    rows truncated" are
  truncated views for your reading; pointing a sentinel 
      at them is still
  valid — the substitution uses the full underlying rows.
- If no execute_result matches the listing the user 
    wants, shape the answer
  normally from the visible rows.

### OUTPUT FORMAT
- Pipe-separated tuples for multiple rows: (v1, v2) | 
    (v1, v2)
- Single tuple for one row with multiple columns: (v1, 
    v2, v3, ...)
- Single value for a scalar: v1
- Sentinel for passthrough: <execute_result_N>
- Use EXACT values from the execute results (do not 
    round or modify numbers).
- Output ONLY the formatted answer, nothing else. No 
    explanation, no markdown.

\end{Verbatim}
\end{promptbox}

\end{document}